\pdfoutput=1

\documentclass[11pt]{article}

\usepackage{coling}

\usepackage{times}
\usepackage{latexsym}

\usepackage[T1]{fontenc}

\usepackage[utf8]{inputenc}

\usepackage{microtype}

\usepackage{inconsolata}

\usepackage{graphicx}
\usepackage{booktabs}
\usepackage{amsmath}
\usepackage{amssymb}
\usepackage{amsfonts}
\usepackage{enumitem}
\usepackage{multirow}
\usepackage{tikz}
\usepackage{pgfplots}
\pgfplotsset{compat=1.5.1}
\def\addlegendimage{\csname pgfplots@addlegendimage\endcsname}
\usetikzlibrary{patterns}
\usetikzlibrary{pgfplots.groupplots}
\usetikzlibrary{
  pgfplots.colorbrewer,
}

\definecolor{purple}{rgb}{0.56, 0.0, 1.0}

\newcommand\blfootnote[1]{%
  \begingroup
  \renewcommand\thefootnote{}\footnote{#1}%
  \addtocounter{footnote}{-1}%
  \endgroup
}

\title{Rethinking KenLM: Good and Bad Model Ensembles \\ for Efficient Text Quality Filtering in Large Web Corpora}

\author{Yungi Kim$^{*}$, Hyunsoo Ha$^{*}$, Sukyung Lee, Jihoo Kim, Seonghoon Yang, Chanjun Park$^{ \dagger}$ \\
\\
  Upstage AI \\
  \texttt{\{eddie, hyunsooha, sukyung, jerry, hoonyang, chanjun.park\}@upstage.ai}}

\begin{document}
\maketitle
\begin{abstract}
\blfootnote{$^*$Equal Contribution $^\dagger$ Corresponding Author}
With the increasing demand for substantial amounts of high-quality data to train large language models (LLMs), efficiently filtering large web corpora has become a critical challenge. For this purpose, KenLM, a lightweight n-gram-based language model that operates on CPUs, is widely used. However, the traditional method of training KenLM utilizes only high-quality data and, consequently, does not \textit{explicitly} learn the linguistic patterns of low-quality data. To address this issue, we propose an ensemble approach that leverages two contrasting KenLMs: (i) \textit{Good KenLM}, trained on high-quality data; and (ii) \textit{Bad KenLM}, trained on low-quality data. Experimental results demonstrate that our approach significantly reduces noisy content while preserving high-quality content compared to the traditional KenLM training method. This indicates that our method can be a practical solution with minimal computational overhead for resource-constrained environments.
\end{abstract}

\section{Introduction}
The advancement of large language models (LLMs) has accelerated as the \textit{`scaling law'}~\cite{kaplan2020scaling}, which states that the performance of LLMs directly correlates with data size, has been studied. Moreover, recent studies~\cite{NEURIPS2023_fa3ed726, gunasekar2023textbooks, li2024datacomp, penedo2024fineweb, dubey2024llama} have shown that the performance of LLMs is largely determined by the quality of the training corpus. In other words, a vast amount of high-quality training corpus is necessary to enhance the performance of LLMs.

However, large web corpora often contain substantial amounts of low-quality data, making them difficult to use directly for training. In response to this challenge, various methods~\cite{pmlr-v235-wettig24a, kong2024large} are employed to filter out low-quality data and select high-quality data. These methods typically require GPU resources, which makes them impractical, especially when processing data that exceeds trillions of tokens.

To \textit{efficiently} filter large datasets, the most widely used method is KenLM~\cite{heafield2011kenlm}, a lightweight n-gram-based model that operates on \textit{CPUs}. In many studies~\cite{wenzek2019ccnet, together2023redpajama, nguyen2023culturax, laurenccon2024obelics}, KenLM, trained on the high-quality Wikipedia dataset, is commonly used. It measures perplexity (PPL) to identify low-quality content. Note that higher PPL scores indicate lower-quality or out-of-domain text, while lower PPL scores suggest that the text closely resembles the linguistic patterns of the high-quality data used to train KenLM. Low-quality data with high PPL scores are then filtered out.

We argue that the traditional KenLM does not \textit{explicitly} learn the linguistic patterns of low-quality data. Thus, while it assigns low PPL scores to data with high-quality linguistic patterns, it does not \textit{consistently} assign high PPL scores to data with low-quality linguistic patterns. To address this issue, we propose an ensemble approach that utilizes the following two contrasting KenLMs: (i) \textit{Good KenLM}, trained on high-quality data; and (ii) \textit{Bad KenLM}, trained on noisy, low-quality data such as spam emails, hate speech, and informal social media text. Our empirical results show that this approach can be a practical solution with minimal computational overhead for resource-constrained environments, significantly reducing noisy content and preserving high-quality content compared to the traditional KenLM training method.

\section{Related Work}
As the demand for a vast amount of high-quality training corpus grows, it has become essential to \textit{effectively} and \textit{efficiently} filter large amounts of web corpus. Among various filtering methods, this paper focuses on model-based quality filtering, which can be broadly divided into the following two categories: (i) perplexity-based filtering; and (ii) classifier-based filtering.

\paragraph{Perplexity-based filtering.}
Numerous studies~\cite{wenzek2019ccnet, together2023redpajama, nguyen2023culturax, wei2023polylm, paster2023openwebmath, laurenccon2024obelics} use the perplexity (PPL) scores of KenLM~\cite{heafield2011kenlm}, an n-gram-based language model, to \textit{efficiently} filter out low-quality data due to its lightweight architecture. It can operate on \textit{CPUs}, making it a cost-efficient solution for handling large-scale text data. Despite its efficiency, there have been few efforts to improve its performance.
Meanwhile, The Pile~\cite{gao2020pile} used the perplexity of GPT-2~\cite{radford2019language} and GPT-3~\cite{brown2020language} to evaluate the quality of the dataset.

\paragraph{Classifier-based filtering.}
FastText~\cite{joulin2016bag} is widely used to distinguish the quality of data~\cite{together2023redpajama, wei2023polylm, li2024datacomp}. 
Similar to KenLM, FastText is also an efficient model that operates on CPUs. However, as detailed in Section~\ref{sec:experiments}, KenLM demonstrated superior performance compared to FastText when both were trained on the same dataset.
Furthermore, recent research~\cite{gunasekar2023textbooks, li2024datacomp, penedo2024fineweb} has focused on fine-tuning pre-trained embedding models to serve as classifiers for quality filtering. Especially, Fineweb demonstrated that training relatively small-sized LLMs (1.82 billion parameters) on data filtered by a trained classifier (resulting in 350 billion tokens), rather than on unfiltered data, led to performance improvements across various benchmarks. However, these methods are impractical for processing large web corpora due to their high computational costs, which necessitate significant GPU resources.

\section{Proposed Method}\label{sec:method}
In this paper, we aim to reduce noisy data while preserving high-quality data in a computationally efficient manner. To this end, we propose an ensemble approach using two contrasting KenLMs: (i) Good KenLM and (ii) Bad KenLM.

\paragraph{Good KenLM.}
The objective of Good KenLM is to assign low perplexity (PPL) scores to well-structured, high-quality text. Many previous studies~\cite{wenzek2019ccnet, together2023redpajama, nguyen2023culturax, laurenccon2024obelics} have used a high-quality Wikipedia dataset for training, denoted as Wiki KenLM in this paper. However, with recent advancements in LLMs, several high-quality datasets~\cite{soldaini-etal-2024-dolma, penedo2024fineweb, li2024datacomp} have emerged. In our experiments, as shown in Section~\ref{sec:experiments}, we found that the combination of S2ORC~\cite{lo-wang-2020-s2orc} and Textbooks-are-all-you-need-lite~\cite{textbooks} as training data was more effective than utilizing Wikipedia. Thus, in this paper, we designate the KenLM trained on this combination of data as Good KenLM.

\paragraph{Bad KenLM.}
The rationale behind employing Bad KenLM alongside Good KenLM is that Good KenLM fails to detect unwanted content (\textit{e.g.}, spam, advertising, and informal communication), which are generally considered poor for training LLMs, as it has not been \textit{explicitly} trained on these types of content. For instance, if low-quality content shares superficial linguistic patterns with high-quality text, it may still score reasonably well under Good KenLM.
Therefore, to detect a wider range of undesirable content, Bad KenLM is designed to assign low PPL scores to such content. To achieve this, we trained Bad KenLM using noisy, low-quality datasets, including hate speech, spam emails, and informal social media content. To the best of our knowledge, this is the first study to employ KenLM trained on noisy, low-quality datasets.

\paragraph{Ensemble.}
To leverage the complementary strengths of two contrasting KenLMs, we ensemble the models by integrating the PPL scores assigned by each. We perform Z-score standardization to align the scales of the two PPL scores assigned by each model, as they are trained on different datasets and therefore exhibit different distributions of PPL scores. Then, we compute the ensembled PPL score $P_{ens}(x_i)$, as follows:

\footnotesize
\begin{equation}
\begin{aligned}
    P_{\text{ens}}(x_i) = & \, \alpha \left( \frac{P_{\text{good}}(x_i) - \mu_{\text{good}}}{\sigma_{\text{good}}} \right) \\
    & - (1 - \alpha) \left( \frac{P_{\text{bad}}(x_i) - \mu_{\text{bad}}}{\sigma_{\text{bad}}} \right),
\end{aligned}
\label{eq:1}
\end{equation}
\normalsize
where $x_i \in \mathcal{X}$ denotes the $i$-th text data, $\mathcal{X}$ represents datasets, $P_{\text{good}}(x_i)$ (resp. $P_{\text{bad}}(x_i)$) indicates PPL score from Good (resp. Bad) KenLM for $x_i$, $\mu_{\text{good}}$ (resp. $\mu_{\text{bad}}$) is the mean of the PPL scores from Good (resp. Bad) KenLM, $\sigma_{\text{good}}$ (resp. $\sigma_{\text{bad}}$) is the standard deviation of the PPL scores from Good (resp. Bad) KenLM, and $\alpha$ denotes a parameter that balances the two PPL scores.
Note that the coefficient for the term associated with Bad KenLM is negative. This is because, in contrast to Good KenLM, which assigns low PPL scores to high-quality data, Bad KenLM assigns low PPL scores to low-quality data. Consequently, data with low ensembled PPL scores—obtained by appropriately subtracting two PPL scores—closely resemble the linguistic patterns of high-quality data and are distinctly separated from low-quality content.

\section{Experiments}\label{sec:experiments}
We designed our experiments to answer the following key research questions (RQs):
\begin{itemize}[leftmargin=*, itemsep=0.1em]
\item \textbf{RQ1}: Does our ensemble approach outperform existing models in removing noisy content while preserving high-quality content?
\item \textbf{RQ2}: Which data sources are effective for training the Bad KenLM?
\item \textbf{RQ3}: How sensitive is the performance of our ensemble approach to hyperparameter $\alpha$?
\item \textbf{RQ4}: How much additional computational overhead does our ensemble approach introduce compared to a single KenLM?
\item \textbf{RQ5}: What types of data does our ensemble approach effectively filter out?
\end{itemize}

\subsection{Experimental Settings}
\paragraph{Dataset and model details.}
As mentioned in Section~\ref{sec:method}, we randomly selected subsets of 300,000 samples each from S2ORC~\cite{lo-wang-2020-s2orc} and Textbooks-are-all-you-need-lite~\cite{textbooks} as training data for Good KenLM. For the training data of Bad KenLM, we collected datasets that is likely to hinder the training of LLMs. Specifically, we used 1,000,000 pieces of social network service (SNS) data (Twitter)~\cite{stockmarkettweets, trumptweets, elontweets, cryptostocktweets} and 776,142 pieces of spam message data~\cite{eronspam, telegramspamham, hamspamscamtoxic}. During the training of both models, we configured the n-gram size to $6$ and the vocabulary size to $65,536$. Also, we set the hyperparameter $\alpha$ to 0.7.

\paragraph{Evaluation details.}
To evaluate the effectiveness of our ensemble approach, we measured perplexity (PPL) scores for the CC-MAIN-2024-10 dump (211 million samples) from Fineweb-edu~\cite{penedo2024fineweb}. Following \citet{wenzek2019ccnet, together2023redpajama}, we then filtered the data based on the 30th and 60th percentiles of PPL scores. Subsequently, we measured the proportion of data with an \textit{educational score} of $2.5$ or higher that was included. In other words, we treated data with an educational score of $2.5$ or higher as the ground truth and measured the recall value. Note that the educational scores are annotated using extensive GPU resources, and it has been demonstrated that training LLMs with data possessing high educational scores leads to performance improvements.

\subsection{Main Results}
We highlight the best results in bold and the second-best results with an underline in the tables.

\paragraph{RQ1: Comparison of existing models.}
As shown in Table~\ref{table1}, our Good KenLM significantly outperformed the widely used Wiki KenLM. Although Bad KenLM alone showed poor performance, our strategy of ensembling it with Good KenLM outperformed even FastText trained on the same data, improving Recall@30 and Recall@60 by 9.76\% and 2.50\%, respectively.

\begin{table}[t]
  \centering
  \small
  \resizebox{0.45\textwidth}{!}{
      \begin{tabular}{c|ccc}
        \toprule
        \textbf{Models}                        & \textbf{Recall@30} & \textbf{Recall@60} & \textbf{Average Recall} \\
        \midrule
        Wiki KenLM                             & 0.5530             & 0.8513             & 0.7022                 \\
        Good KenLM                             & 0.7059             & 0.9195             & 0.8127                 \\ 
        Bad KenLM                              & 0.3403             & 0.7031             & 0.5217                 \\ 
        \midrule
        FastText(Wiki, Bad)                    & 0.6453             & 0.8878             & 0.7665                 \\
        FastText(Good, Bad)                    & 0.7462             & 0.9412             & 0.8437                 \\ 
        \midrule
        Ens(Good, Bad)                         & \textbf{0.8190}    & \textbf{0.9647}    & \textbf{0.8919}        \\
        \midrule\midrule
        Ens(Good, Wiki)                        & 0.6312             & 0.8898             & 0.7605                 \\
        \bottomrule
    \end{tabular}
    }
    \vspace{-0.2cm}
  \caption{
    Performance comparison of our approach with existing models, and an ablation study on our design choices.
  }
  \vspace{-0.3cm}
  \label{table1}
\end{table}

Moreover, to validate the effectiveness of Bad KenLM within our ensemble framework, we conducted a comparative experiment where Good KenLM and Wiki KenLM were ensembled in place of Bad KenLM, denoted as \textit{Ens(Good, Wiki)}. The performance of \textit{Ens(Good, Wiki)} was lower than that of Good KenLM alone. This is likely due to the relatively lower quality of the Wikipedia dataset compared to the training data used for Good KenLM, which negatively impacts its overall performance. This result also highlight the importance of incorporating Bad KenLM into the ensemble, as it successfully identifies undesirable content that Good KenLM may overlook.

\paragraph{RQ2: Impact of data sources on training Bad KenLM.}
The training dataset for Bad KenLM is diverse, including SNS, spam mail, and toxic datasets~\cite{hateoffensive, gibert2018hate, kennedy2020constructing, mathew2020hatexplain, vidgen2021learningworstdynamicallygenerated, pavlopoulos-etal-2022-acl} containing hate speech and profanity. We conducted experiments to determine which of these data sources are effective for training Bad KenLM.
In this experiment, we ensembled our Good KenLM with various Bad KenLMs, each trained on different combinations of datasets.

As shown in Table~\ref{table2}, SNS data (Twitter) proved to be the most effective for training Bad KenLM, which is designed to filter out noisy content unsuitable for LLM training. Interestingly, toxic datasets led to a decrease in the performance of Bad KenLM. Unlike SNS data or spam mail, which share similar distributions with web data, toxic datasets contain a large proportion of highly offensive language, resulting in a substantial distributional difference. This discrepancy seems to adversely affect the training process of Bad KenLM.
\begin{table}[t]
  \centering
  \small
  \resizebox{0.45\textwidth}{!}{
      \begin{tabular}{c|ccc}
        \toprule
        \multirow{2}{*}{\shortstack{\textbf{Training Dataset}\\\textbf{of Bad KenLM}}} & \multicolumn{3}{c}{\textbf{Metrics}} \\
        & \textbf{Recall@30} & \textbf{Recall@60} & \textbf{Average Recall} \\
        \midrule
        Spam                             & 0.8059             & 0.9576             & 0.8818                 \\
        Twitter                          & \underline{0.8131} & \textbf{0.9651}    & \underline{0.8891}     \\
        Toxic                            & 0.7320             & 0.9402             & 0.8361                 \\
        Spam + Twitter                   & \textbf{0.8190}    & \underline{0.9647} & \textbf{0.8919}        \\
        Spam + Toxic                     & 0.7885             & 0.9545             & 0.8715                 \\
        Twitter + Toxic                  & 0.7973             & 0.9602             & 0.8788                 \\
        Spam + Twitter + Toxic           & 0.7906             & 0.9533             & 0.8720                 \\
        \bottomrule
    \end{tabular}
  }
  \vspace{-0.2cm}
  \caption{
    The effect of data sources on Bad KenLM training.
  }
  \vspace{-0.3cm}
  \label{table2}
\end{table}

\paragraph{RQ3: Hyperparameter sensitivity analysis.}
The parameter $\alpha$ in Eq.~\eqref{eq:1} adjusts the balance between the PPL scores of Good KenLM and Bad KenLM. We analyze how the performance of our ensemble approach varies with changes in $\alpha$ in terms of Recall@30 and Recall@60.

\begin{figure}[t]
\centering
\begin{tikzpicture}
\scriptsize
\begin{axis}[
height=2.8cm,
width=4.5cm,
xtick={1, 2, 3, 4, 5, 6, 7, 8, 9, 10, 11},
xticklabels={0, 0.1, 0.2, 0.3, 0.4, 0.5, 0.6, 0.7, 0.8, 0.9, 1},
ylabel=Recall@30 (\%),xlabel=$\alpha$,
ymin=0.20, ymax=1.0,
xticklabel style={rotate=50, font=\tiny},
x label style={at={(0.5,-0.35)}},
y label style={at={(-0.15,0.5)}},
y tick label style={/pgf/number format/.cd,fixed,fixed zerofill,precision=1,/tikz/.cd,font=\tiny}]
\addplot[color=purple,mark=square,]
coordinates {(1, 0.3403) (2, 0.3756) (3, 0.4250) (4, 0.4948) (5, 0.5906) (6, 0.7056) (7, 0.8003) (8, 0.8190) (9, 0.7921) (10, 0.7506) (11, 0.7059)};
\end{axis}
\end{tikzpicture}
\begin{tikzpicture}
\scriptsize
\begin{axis}[
height=2.8cm,
width=4.5cm,
xtick={1, 2, 3, 4, 5, 6, 7, 8, 9, 10, 11},
xticklabels={0, 0.1, 0.2, 0.3, 0.4, 0.5, 0.6, 0.7, 0.8, 0.9, 1},
ylabel=Recall@60 (\%),xlabel=$\alpha$,
ymin=0.5, ymax=1.1,
xticklabel style={rotate=50, font=\tiny},
x label style={at={(0.5,-0.35)}},
y label style={at={(-0.15,0.5)}},
y tick label style={/pgf/number format/.cd,fixed,fixed zerofill,precision=1,/tikz/.cd,font=\tiny}]
\addplot[color=purple,mark=square,]
coordinates {(1, 0.7031) (2, 0.7395) (3, 0.7872) (4, 0.8476) (5, 0.9149) (6, 0.9620) (7, 0.9711) (8, 0.9647) (9, 0.9520) (10, 0.9362) (11, 0.9195)};
\end{axis}
\end{tikzpicture}
\vspace{-0.2cm}
\caption{The effect of $\alpha$ on the performance of our ensemble approach.}\label{fig:alpha}
\vspace{-0.5cm}
\end{figure}
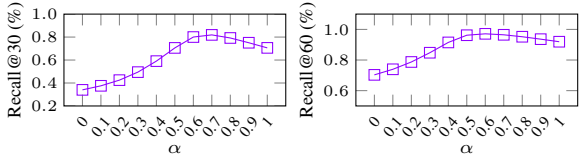

As depicted in Figure~\ref{fig:alpha}, Recall@30 and Recall@60 continuously improve as $\alpha$ increases to 0.7 and 0.6, respectively, and then gradually decrease. These results suggest that when $\alpha$ is too small, the influence of Bad KenLM becomes overly dominant, resulting in poor preservation of high-quality content. Conversely, when $\alpha$ is too large, the influence of Good KenLM prevails, leading to the inclusion of some low-quality content. These results indicate that appropriately determining the value of $\alpha$ is critical for effectively removing noisy content while preserving high-quality content.

\paragraph{RQ4: Degree of computational overhead.}
To assess the computational overhead of our approach, we measured the processing time and estimated cost\footnote{It was measured using an AWS r6a.32xlarge~\cite{AWSR6AInstances} spot instance.} for the CC-MAIN-2024-10 dump on a machine with 128-core CPUs. As presented in Table~\ref{table3}, our approach increased the processing time from 2,234 to 3,928 seconds, with an additional cost of \$1.08. These increases are justified by the recall improvement from 81.27\% to 89.19\%, as high-quality data is crucial for effective LLM training.

\begin{table}[t]
  \centering
  \resizebox{0.48\textwidth}{!}{
      \begin{tabular}{c|cccc}
        \toprule
        \textbf{Models} & \textbf{Processing Time} & \textbf{Estimated Cost} & \textbf{Throughput} & \textbf{Avg. Recall}\\
        \midrule
        Good KenLM & 2,234s & \$1.42 & 94.4k docs/s & 0.8127 \\
        Ens(Good, Bad) & 3,928s & \$2.50 & 53.7k docs/s & 0.8919 \\
        \bottomrule
    \end{tabular}
  }
  \vspace{-0.2cm}
  \caption{
    Comparison of computational overhead and performance for the CC-MAIN-2024-10 dump between Good KenLM and our ensemble approach.
  }
  \vspace{-0.2cm}
  \label{table3}
\end{table}

\begin{figure}[t]
\centering
\includegraphics[width=0.45\textwidth]{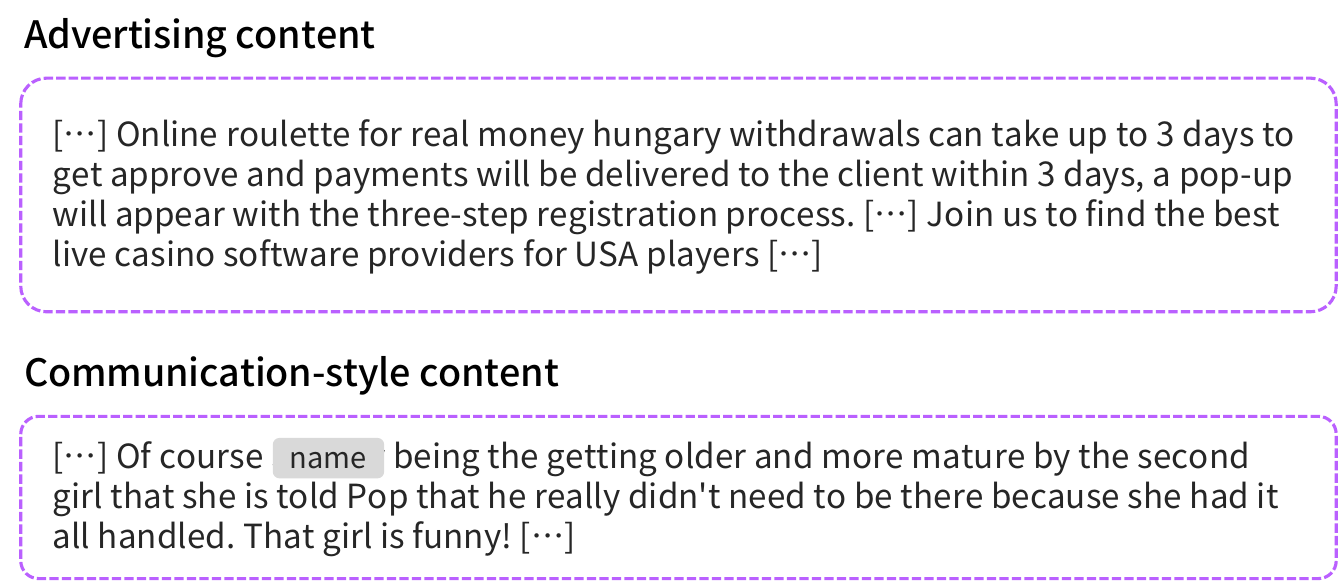}
\vspace{-0.1cm}
\caption{Visualization of examples that are not filtered by Good KenLM but are successfully removed by our ensemble approach.} \label{fig:case}
\vspace{-0.5cm}
\end{figure}

\paragraph{RQ5: Case study on the effectiveness of our approach.}
To demonstrate the effectiveness of our ensemble approach, we present examples that are not filtered by Good KenLM but are successfully removed by our ensemble approach. As illustrated in Figure~\ref{fig:case}, our approach effectively filters advertising and communication-style content, which are generally unsuitable for LLM training. 
Since advertising content is usually written politely, Good KenLM, trained only on high-quality datasets, struggles to detect it. Conversely, Bad KenLM, trained on spam mail and SNS data, successfully identifies such content as well as communication-style content. Therefore, our ensemble approach more effectively filters these types of content.

\section{Conclusion}
In this paper, we propose an ensemble approach using Good KenLM and Bad KenLM for effective text filtering. By integrating perplexity scores, we successfully filter out noisy data, such as spam and informal content, while preserving high-quality text. Empirical results suggest that our approach could be a practical solution for filtering large-scale datasets in resource-constrained environments.

\section*{Acknowledgments}
We would like to thank the members of Upstage and our collaborators for their valuable feedback and support throughout the development of this work. 

This work was supported by Institute of Information \& Communications Technology Planning \& Evaluation(IITP) grant funded by the Korea government(MSIT) (No. RS-2024-00338140, Development of learning and utilization technology to reflect sustainability of generative language models and up-to-dateness over time).

\section*{Limitations}
While the proposed method using Good KenLM and Bad KenLM offers effective filtering of large-scale datasets, it has the following limitations: (i) Although our method has demonstrated effectiveness through extensive experiments using Fineweb-edu, we have not been able to measure its direct impact on LLMs training due to computational cost constraints; and (ii) the model relies heavily on predefined training datasets, and its performance may degrade when applied to content that significantly differs from the training corpora.

\section*{Ethics Statement}
The experiments conducted in this paper were carried out objectively and fairly. No biases were introduced during the data selection or evaluation process. All datasets used in this research are publicly available, and the methods were rigorously tested to ensure the reliability and validity of the results.

\bibliography{custom}

\appendix

\end{document}